\documentclass[preprint]{vgtc}               

\ifpdf
  \pdfoutput=1\relax                   
  \usepackage{graphicx}                
  \DeclareGraphicsExtensions{.pdf,.png,.jpg,.jpeg} 
\else
  \usepackage{graphicx}                
  \DeclareGraphicsExtensions{.eps}     
\fi%

\graphicspath{{figures/}{pictures/}{images/}{./}} 

\usepackage{microtype}                 
\PassOptionsToPackage{warn}{textcomp}  
\usepackage{textcomp}                  
\usepackage{mathptmx}                  
\usepackage{times}                     
\usepackage{cite}                      
\usepackage{tabu}                      
\usepackage{booktabs}                  

\onlineid{0}

\vgtccategory{Research}

\vgtcinsertpkg

\preprinttext{IEEE VR 2019 Workshop on Human Augmentation and its Applications}

\title{Post-Data Augmentation to Improve Deep Pose Estimation of Extreme and Wild Motions}

\author{Kohei Toyoda\thanks{e-mail: toyoda-kohei648@g.ecc.u-tokyo.ac.jp}\\ %
        \scriptsize The University of Tokyo %
\and Michinari Kono\thanks{e-mail: mchkono@acm.org}\\ %
     \scriptsize The University of Tokyo %
\and Jun Rekimoto\thanks{e-mail: rekimoto@acm.org}\\ %
     \parbox{1.4in}{\scriptsize \centering The University of Tokyo \\ Sony Computer Science Laboratories, Inc.}}

\teaser{
 \includegraphics[width=1.0\textwidth]{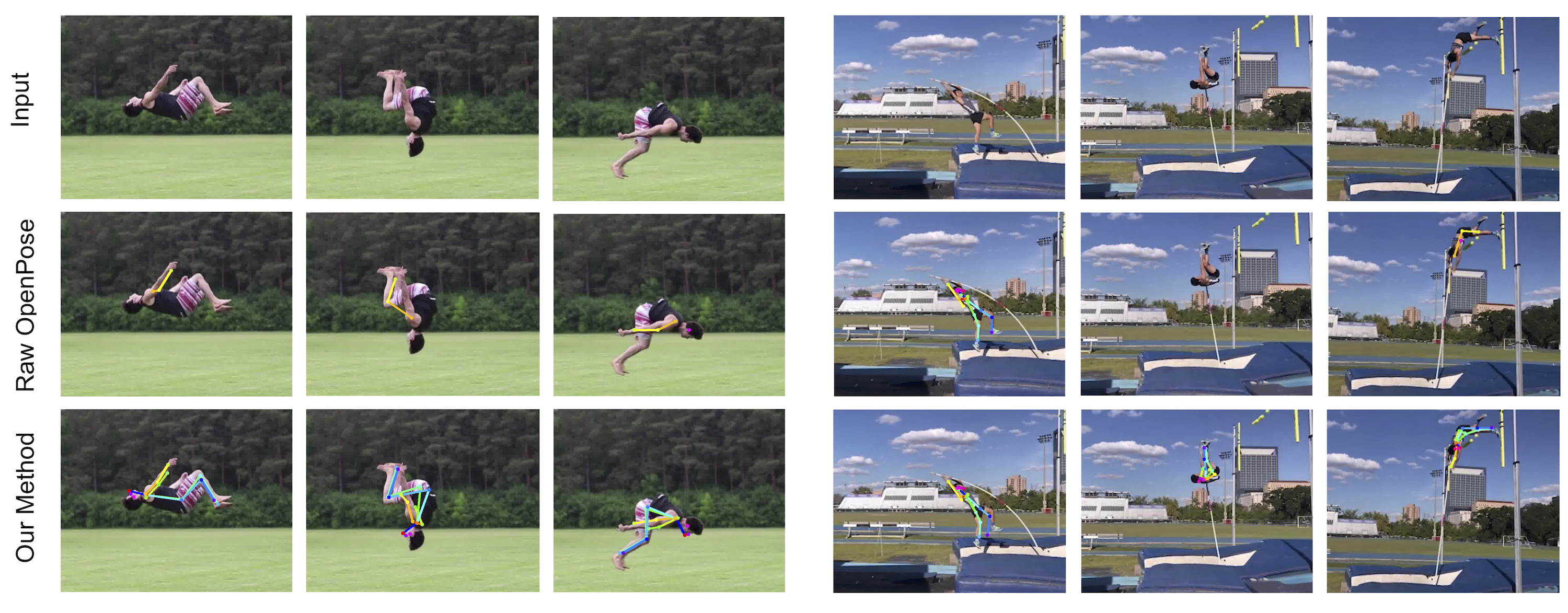}
 \caption{Applying data augmentation approach for the estimation/generation phase in deep pose estimation can improve the quality of extreme/wild motion pose estimation. Our approach can be used with pre-trained models, and without new training with self-collected dataset. This figure shows results compared with using raw OpenPose~\cite{cmupose}.}
  \label{fig:rep}
}

\abstract{Contributions of recent deep-neural-network (DNN) based techniques have been playing a significant role in human-computer interaction (HCI) and user interface (UI) domains. One of the commonly used DNNs is human pose estimation. This kind of technique is widely used for motion capturing of humans, and to generate or modify virtual avatars. However, in order to gain accuracy and to use such systems, large and precise datasets are required for the machine learning (ML) procedure. This can be especially difficult for extreme/wild motions such as acrobatic movements or motions in specific sports, which are difficult to estimate in typically provided training models. In addition, training may take a long duration, and will require a high-grade GPU for sufficient speed. To address these issues, we propose a method to improve the pose estimation accuracy for extreme/wild motions by using pre-trained models, i.e., without performing the training procedure by yourselves. We assume our method to encourage usage of these DNN techniques for users in application areas that are out of the ML field, and to help users without high-end computers to apply them for personal and end use cases.%
} 

\CCScatlist{
  \CCScatTwelve{Computing methodologies}{Motion capture}{}{};
  \CCScatTwelve{Computing methodologies}{Neural networks}{}{};
  \CCScatTwelve{Human-centered computing}{Human computer interaction (HCI)}{}{}
}

\begin{document}


\firstsection{Introduction}

\maketitle


\begin{figure*}[]
 \centering 
 \includegraphics[width=0.9\textwidth]{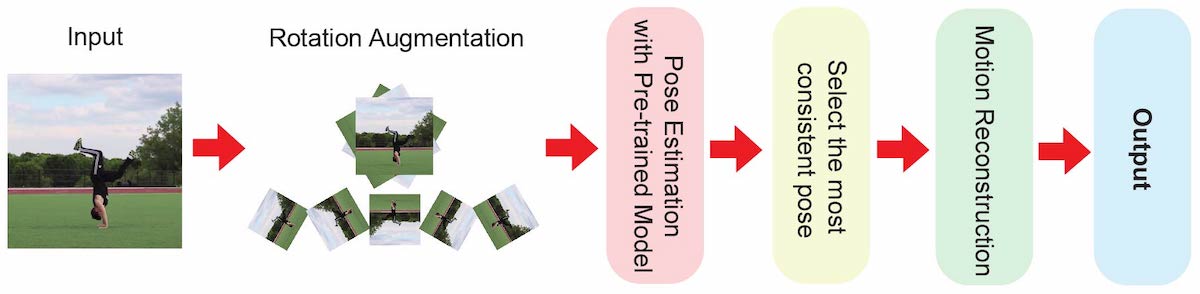}
 \caption{The overview of our approach. Instead of augmenting data for training, we apply augmentation for the estimation process. Pose estimation is applied multiple times for each frame, and then we select the best consistent pose from the set. A simple reconstruction is applied to get the final output.}
 \label{fig:overview}
\end{figure*}

\begin{figure*}[]
 \centering 
 \includegraphics[width=1.0\textwidth]{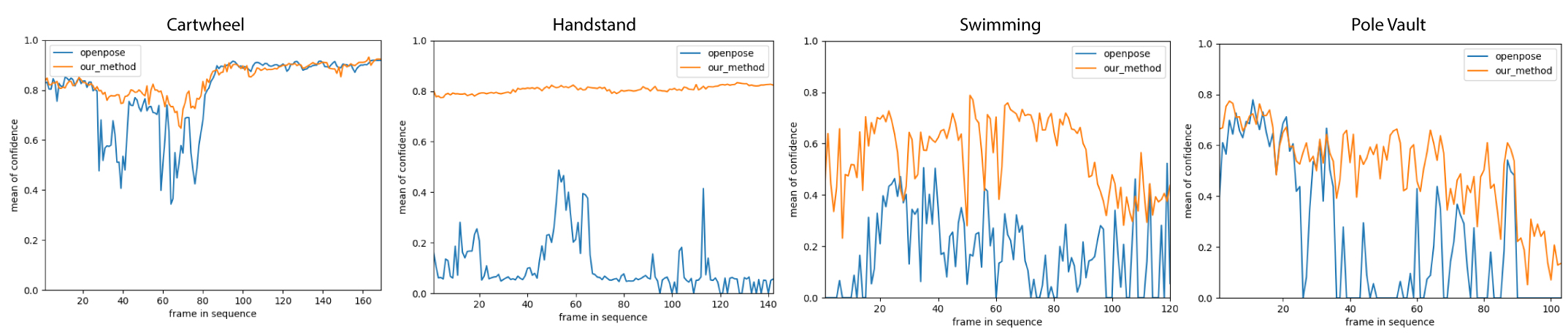}
 \caption{Confidence levels of some example video sequences. Note that the confidence do not exactly present the accuracy, but is a simple reference for the quality of prediction. The confidence level presented here is the output retrieved from OpenPose, with and without augmentation with our method.}
 \label{fig:conf}
\end{figure*}

Measuring human pose is an important topic for analyzing the motion of the human body, which has been studied and applied to various research fields, such as sports science and human-computer interaction (HCI). Typical approaches for motion capturing were to use optical or mechanical sensors fixed or worn by the user. However, the recent growth of the deep-neural-network (DNN) techniques have presented remarkable results for human pose estimation by using simple monocular cameras that can retrieve 2D poses ~\cite{cmupose,cnnpose,deeppose} or even 3D poses~\cite{kanazawahmr,3dposebaseline}. These approaches have several benefits, for example, they do not require camera calibration and can be obtained by single RGB images. 

In DNN techniques, they usually require to have large datasets that can be used for training models, and online available data is used (e.g.,~\cite{h36m_pami}). Unfortunately, using such data sometimes find difficulty to be applied for rare or specific postures, such as extreme or wild motions. In order to solve this problem, a typical solution is to gather data by yourselves by using other motion capturing tools for reference data, however, there are still difficulties to gather data for some motions. For example, gathering data for swimming may require an expensive or special environment for motion capturing. Instead of gathering such data, Peng et al.~\cite{pengsfv} suggested that rotation data augmentation can improve the performance of the pose estimation for {\it in the wild} images. This solution may improve the quality of pose estimation for extreme and wild motions. 

However, this may result to another problem, where the training data becomes very large. Training with large dataset may require much more time for training and require high-grade GPUs. Therefore, it may be difficult for some {\it users} to use the DNN pose estimation by themselves. For such {\it users}, they may find benefit from just using the pre-trained model that is often provided by the developers. What we mean by saying {\it users} here, are people working out of the ML field, who do not develop DNN by themselves, but use them for developing applications. 

DNN pose estimation techniques are used for various applications, for both realtime and non-realtime use cases. Move Mirror\footnote{https://experiments.withgoogle.com/collection/ai/move-mirror/view} allows you to find images with the similar pose with the input. Other applications include use cases for entertainment purposes~\cite{dancevideo,parapara} and sport analysis~\cite{swimpose,freethrowpose,bodyshots}. Some even focus on analyzing hands or facial expressions for other purposes~\cite{cookpose,syncclass}. We can see that pose estimation is important for the users, and do not always have to be available for realtime use.

In this paper, we propose a simple but effective method to improve the pose estimation for extreme/wild motion videos, without gathering new reference data, and by just using the provided pre-trained model. Referring to the approach of Peng et al.~\cite{pengsfv}, we apply pose estimation for videos rotated multiple times for each frame to find a confident pose, i.e., we apply a kind of data augmentation technique at the estimation process. As a proof-of-concept, we present example results using OpenPose~\cite{cmupose} with {\it in the wild} videos and some videos recorded in specific activities (Figure~\ref{fig:rep}).

\section{Method}
The approach of our method is shown in Figure~\ref{fig:overview}. First, we apply OpenPose at various rotations to get predictions obtained from various angles, which is expressed by equation~\ref{eq:rotate} and ~\ref{eq:pose}.

\begin{eqnarray}
\label{eq:rotate}
^kj_t^\theta &=& ^kj_t'R^{-1} \\
\label{eq:pose}
J_t^\theta &=& \{^kj_t^\theta\}
\end{eqnarray}

$ ^kj_t'$ is the predicted $k$ th joint at frame $t$ when the image is rotated $\theta$, and $^kj_t^\theta$ is the joint fixed to the original coordinates. $R$ is the rotation matrix for a 2D rotation at $\theta$, and $\theta$ takes values at $\lbrack0^\circ,360^\circ\rbrack$ sampled every $d$ degrees. $d$ is $10^\circ$ in our current case. $J_t^\theta$ is sets of estimated joints when the image is rotated $\theta$. Therefore, $J_t^\theta$ can be understood as a pose that is predicted at a certain rotation but is fixed to the original image coordinates and consists of a set of poses as $\{^kj_t^\theta\}$. When $ t=1$, $J_1^\theta$ takes the pose that satisfies objective function~\ref{eq:max}, where $\bar{x}_t^\theta$ is the mean of confidences of $J_t^\theta$. We then select the ideal $J_t^\theta$ from the set, by using objective function~\ref{eq:objective}. We find the top 5 poses for objective~\ref{eq:objective}, and then use objective~\ref{eq:max} to select the best matching pose. In addition, we defined a threshold for objective~\ref{eq:objective} as 500, and when all the calculation exceeds the threshold, we used objective~\ref{eq:max} to find the best pose instead. This approach was taken because we experimentally found that simply selecting the pose with the best confidence level did not always present the correct pose.


\begin{eqnarray}
\label{eq:max}
&\max \bar{x}_t^\theta \\
\label{eq:objective}
&\displaystyle \min \sum_k \|^kj_t^\theta - ^kj_{t-1}^p \|_2
\end{eqnarray}

Here, $p$ is the final reconstructed pose that is obtained by equation~\ref{eq:final} and~\ref{eq:posed}.
\begin{eqnarray}
\label{eq:final}
J_t^p &=& wJ_t^\theta + (1-w)J_{t-1}^p \\
\label{eq:posed}
J_t^p &=& \{^kj_t^p\}
\end{eqnarray}
$w$ represents the weight for the current frame and is set to $w = 0.8$. $J_t^p$ is used for representing the pose at every $t$.
Note that for the procedures that are taken when selecting the best joints, we do not consider the joints around the head. 



\begin{figure*}[]
 \centering 
 \includegraphics[width=0.9\textwidth]{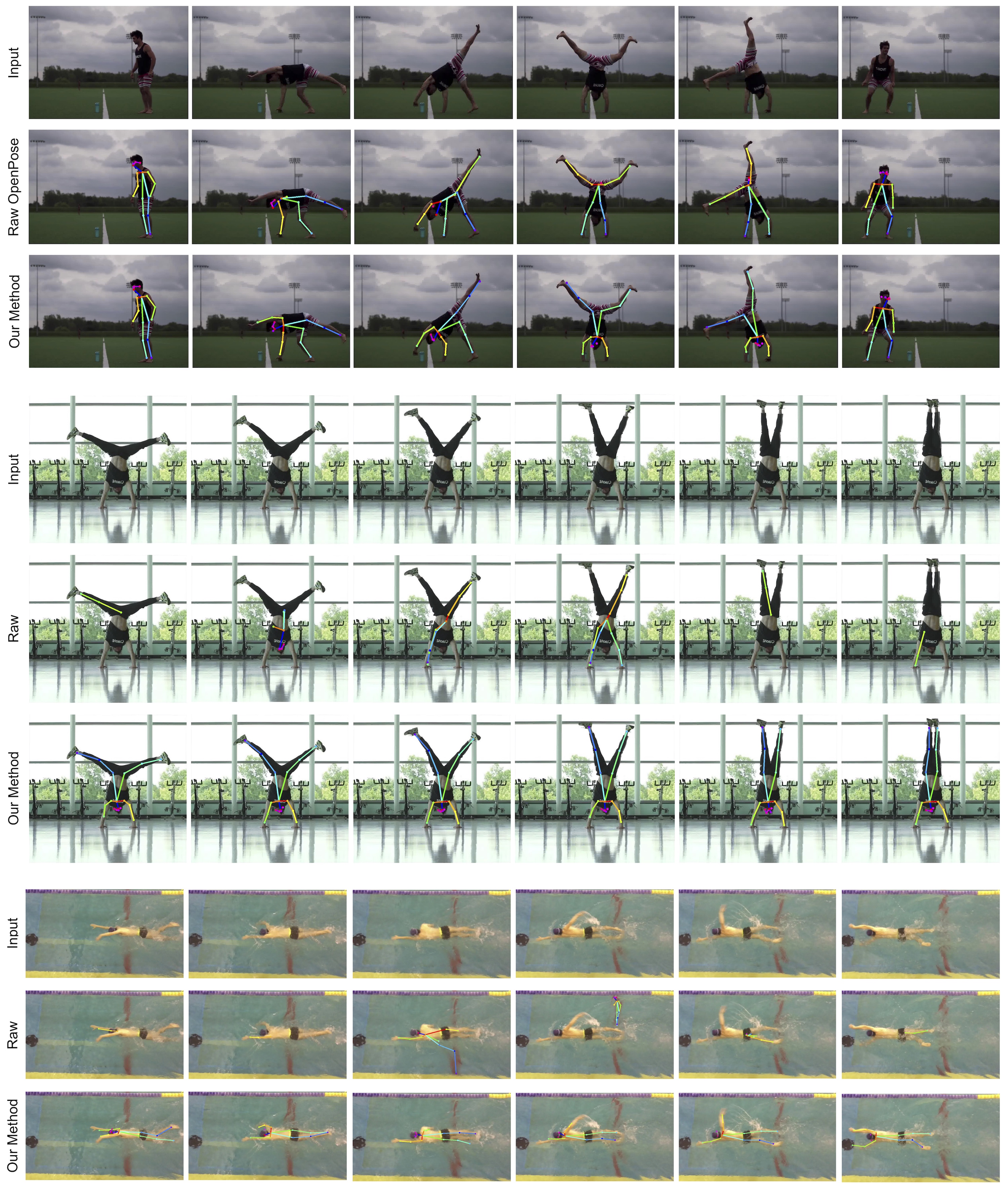}
 \caption{Example result of pose estimation applied to a video sequence. The top row is the input, the middle is the result obtained by using raw OpenPose, and the bottom is the result obtained by using our method.}
 \label{fig:result}
\end{figure*}

\section{Results}
We applied our method for various video sequences, and the followings are some examples.
\begin{itemize}
    \item Cartwheel: A person performs a cartwheel, and jumps vertically at the end.
    \item Handstand: A person performs a handstand, and moves his legs.
    \item Swimming: A person swims across the camera settled above the swimming pool.
    \item Pole Vault: A person performs a pole vault.
\end{itemize}

Figure~\ref{fig:conf} displays confidence levels of some video sequences that were tested. We can observe that the overall confidence level tends to be higher by using our method. However, since we work on typical body models, it is still difficult to estimate poses that exceed the constraints of such models. We can also observe that the confidence level sometimes becomes lower than the results of using raw OpenPose. This happens when the predicted pose has fine confidence, however, do not match the pose from the prior frame, and therefore exceeds the threshold of objective~\ref{eq:objective}.  

Figure~\ref{fig:rep} and~\ref{fig:result} presents some visual results of pose estimation using our method for videos with extreme/wild motions. We can see that using OpenPose with its default usage can suffer to estimate sufficient results especially when the human body is sideways or upside down. Our method successfully allows estimating a better pose throughout the motion. However, we still observed some difficulties for estimating poses of pole vault sequences. This happens when the body curls up, or when the pole disturbs the pose estimation.  

Since our interest was to work on extreme/wild motions that do not have open source dataset provided ({\it in the wild scenes}), we currently have not conducted a quantitative evaluation for the accuracy. In the future, we plan to collect data for these extreme/wild motions with motion captures, and to evaluate our method against other methods, including training models created with the data.

\section{Discussion and Future Work}
Our results suggest that our method can improve the quality of deep pose estimation, however, still has some limitations.

Since our work applies pose estimation multiple times for each frames, it requires more time for the whole pose estimation procedure. Therefore, we can say that there is a trade-off between time and accuracy. We must note that our method cannot be applied for use cases that require realtime analysis. In addition, while the strength of OpenPose is where it can be applied for multi-person simultaneously, our current method focuses on a single person.
     
\begin{figure}[t]
 \centering 
 \includegraphics[width=1.0\columnwidth]{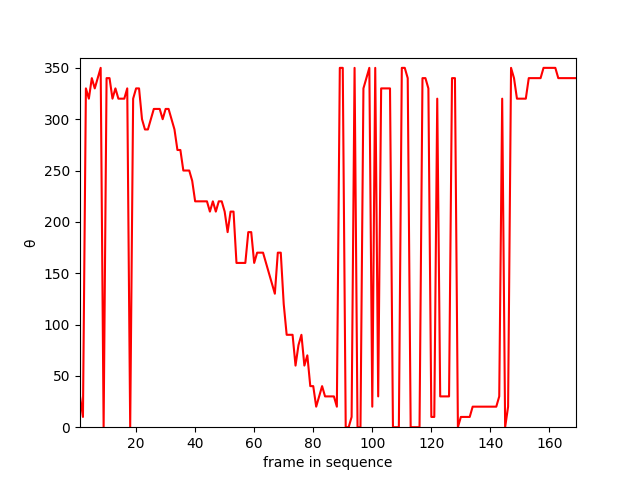}
 \caption{Plot of the $\theta$ value of a video sequence (cartwheel).}
 \label{fig:angle}
\end{figure}

For future work, we will apply our approach to 3D pose estimation as well. We will also investigate other data augmentation approaches to be used for our method. Currently, we apply the pose estimation for all considerable rotation of the frame. However, we assume that the computation cost can be minimizing by referring to the selected rotation angle of adjunct frames, and to limit the rotation angle of the frame so that we can decrease the number of the pose estimation times applied for each frame. Figure~\ref{fig:angle} presents the $\theta$ value of a cartwheel sequence. In this sequence, a person starts standing, and then performs a cartwheel, and finally when the cartwheel is completed, he performs a small jump vertically. During the cartwheel, the $\theta$ value gradually decreases following the progress of the cartwheel. When the person is standing or jumping, $\theta$ takes values near the original angle (typically around 0--30$^\circ$ and 330--360$^\circ$). We may observe that pose estimation can perform better when the head is in the upper area of the image and the legs are at the bottom. Therefore, we may possibly estimate a range of $\theta$, which can make a good estimation.

We also expect our approach of {\it post-data augmentation} can be applied for other application domains, not limited to pose estimation. The approach where we adapt the input data and to help the training model to perform the prediction more easily can improve the overall performance of estimation, and can be effective for various models. We assume that the approach can be expanded with other data augmentation methods, not limited to rotation augmentation. 

Our motivation for this work is to encourage researchers in human augmentation and HCI to use DNN based pose estimation methods for various sports/activities with irregular motions. Introducing a method that does not require data collection and training on your own, will hopefully improve the acceptability and reduce the obstacles for applying such methods.

\section{Conclusion}
This paper introduces a method to improve pose estimation with DNN, but with a post-data augmentation approach that can be applied for pre-trained models. When using traditional methods, it can be difficult to obtain accurate poses from extreme/wild motions. Our method augments the input data with rotation augmentation, and use pose estimation method multiple times for every frame. We then select the most consistent pose, followed by a motion reconstruction for smoothing. The results show that our method can improve the overall quality of pose estimation for videos where people take irregular poses, such as when performing acrobatic motions. We expect our method to be used for applications where collecting datasets for the activity are difficult, and do not require real-time estimation. This approach can be considered as a {\it post-data augmentation} method, which can be used after the training process.

\acknowledgments{
The work is supported by ISID Technosolutions, Ltd.}

\bibliographystyle{abbrv-doi}

\bibliography{template}
\end{document}